\pdfoutput=1

\documentclass[11pt]{article}

\usepackage[preprint]{acl}

\usepackage{times}
\usepackage{latexsym}
\usepackage{todonotes}
\usepackage[T1]{fontenc}

\usepackage[utf8]{inputenc}

\usepackage{microtype}

\usepackage{inconsolata}

\usepackage{graphicx}

\usepackage{booktabs}

\usepackage{authblk}

\newcommand{\verde}[0]{AI-VERDE }

\title{\verde: A Gateway for Egalitarian Access to Large Language Model-Based Resources For Educational Institutions}

\author{\textbf{Paul Mithun}, \textbf{Enrique Noriega-Atala}, \textbf{Nirav Merchant}, {\and}\textbf{Edwin Skidmore} \\
        Data Science Institute \\  University of Arizona \\ \{mithunpaul,enoriega,nirav\}@arizona.edu}

\begin{document}
\maketitle
\begin{abstract}
We present \verde, a unified LLM-as-a-platform service designed to facilitate seamless integration of commercial, cloud-hosted, and on-premise open LLMs in academic settings. \verde streamlines access management for instructional and research groups by providing features such as robust access control, privacy-preserving mechanisms, native Retrieval-Augmented Generation (RAG) support, budget management for third-party LLM services, and both a conversational web interface and API access. In a pilot deployment at a large public university, \verde demonstrated significant engagement across diverse educational and research groups, enabling activities that would typically require substantial budgets for commercial LLM services with limited user and team management capabilities. To the best of our knowledge, AI-Verde is the first platform to address both academic and research needs for LLMs within an higher education institutional framework.

\end{abstract}

\begin{figure*}
    \centering
    \includegraphics[width=0.8\linewidth]{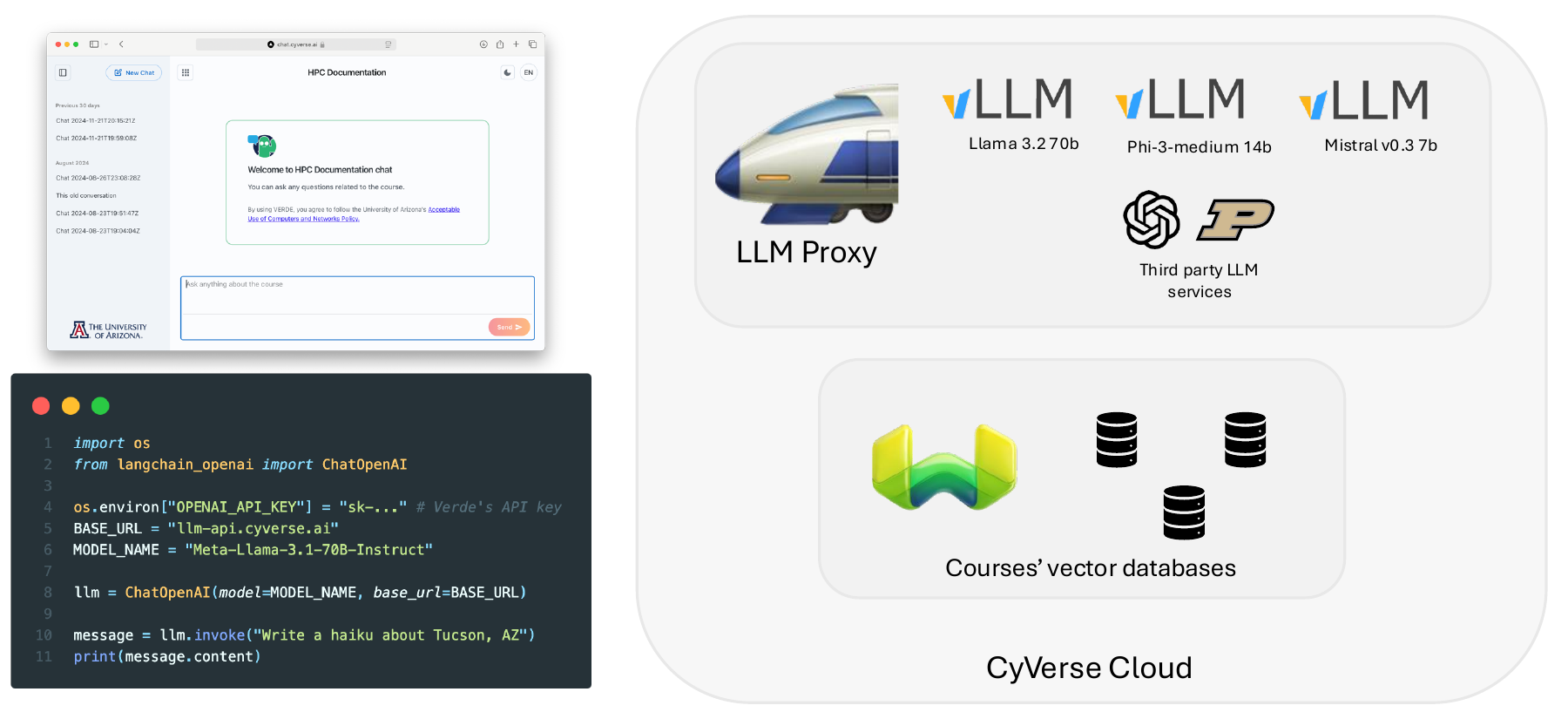}
    \caption{Architecture Diagram of \verde. 
    The left hand side of the diagram represents the \emph{frontend}, which consists of the conversational web interface, depicted at the top, and a snippet of code with an example of how to programmatically connect to \verde using industry-standard python software packages.
    The right side depicts the \emph{backend} elements, and illustrates multiple different models running with vLLM, as well as a proxy to commercial models, all exposed to clients through LiteLLM. The backend also contains our managed instance of the Weaviate vector database manager, which houses the different vector databases, corresponding to each course, enabled in \verde. }
    \label{fig:verde-architecture}
\end{figure*}

\section{Introduction}
Large language models (LLMs) such as ChatGPT \cite{openai2022chatgpt}, GPT-4 \cite{openai2023gpt4}, Mistral \cite{jiang2023mistral}, Llama \cite{dubey2024llama}, etc. have rapidly emerged as transformative tools that demonstrate significant capabilities across a broad spectrum of applications, including natural language processing, content generation, and educational support.  Their powerful capabilities have captivated the attention of users and has led to a huge creative exploration and novel applications that span most disciplines.


Due to this accelerated advance and acceptance of the LLM technology by the hoi polloi, Universities and colleges are also currently under pressure to integrate this into academic settings. However, integrating LLM technology into academic settings, especially higher education institutions, faces several unique challenges, such as, the required  technological know-how, privacy concerns, limited access to specialized knowledge, intellectual property rights etc. For example, a Professor aiming to incorporate commercial LLMs like ChatGPT into their course work often encounters high costs, as well as difficulties in managing usage across students and generating access tokens. Even for using free open-source alternatives like Llama, technical barriers such as programming knowledge and hosting tools are significant, especially for faculty coming from a non-STEM background. Furthermore, even with basic programming knowledge, using an LLM effectively for advanced courses requires fine-tuning or setting up a Retrieval Augmented Generation (RAG) \citep{NEURIPS2020_6b493230} pipeline, both of which involve complex software and hardware requirements. Moreover, issues related to intellectual property, such as the potential public dissemination of copyrighted textbooks, add further challenges to seamless integration.

In this context these are the unique contributions of our work:


    \hspace{-1em}1) We present \verde\footnote{Video demo: \url{https://youtu.be/hPTiZlcUiZo}}, a first of its kind LLM-platform-as-a-service (LLMPaaS) solution which is already being incorporated by several professors and researchers at the University of Arizona.
    
     \hspace{-1em}2)  We present a detailed survey of several problems  that prevent smooth adoption of LLM based technology at Universities and other higher education institutions and show how \verde addresses these concerns.

    \hspace{-1em}3) We do a comparative study with other commercial options and show that AI Verde provides a much lower cost egalitarian gateway to AI tools, and can thus democratize the access to LLMs in colleges and University campuses.
    
    
To the best of our knowledge this is the 
first time, that a survey of such limitations is being done, and a product addressing these limitations, and catering to higher education institutions to holistically address these limitations, as an integrated platform, is being done.

\section{\verde Architecture components}
\label{arch}

The primary goal in designing of \verde was to create egalitarian gateway for members of academia, to smoothly access all facets of AI technology. Hence, while at the core \verde comprises of a standard RAG pipeline to LLMs,
each individual sub-modules and components were made available to be accessed independently as micro-services, in a plug-and-play format.

Specifically, \verde \footnote{\url{https://chat.cyverse.ai/home}} is a platform designed with the goal of providing seamless access to LLMs (especially RAG if need be) to the academic community through various means like chat interfaces, API etc. To achieve this functional requirement, we tapped into open source technologies to serve LLMs as well as developed custom software to allow the provisioning of a multi-tenant environment to accommodate diverse educational and research groups. In this section we detail several of the major architecture components of \verde which is also shown in \autoref{fig:verde-architecture}. The individual pieces are deployed and orchestrated with Kubernetes \cite{Kubernetes2014}.


\subsection{Backend: LLM serving}
At the heart of \verde  lies the LLMs to be exposed to our community. We use vLLM \citep{kwon2023efficient} to persistently load LLM in a GPU cluster. vLLM allows us to serve open source models, such as Llama 3.2 \cite{touvron2023llama2openfoundation}, Mistral \citep{jiang2023mistral7b} and Phi-3 \citep{abdin2024phi} and provide an API interface. Additionally, we leverage the advanced capabilities offered such as support for numerous model weight formats, integration with the HuggingFace Hub\footnote{\url{https://huggingface.co/models}}, support for LoRA adapters \citep{hu2022lora}, and customized paged attention for increased throughput. Each LLM served through \verde is associated with a running instance of vLLM.

We use LiteLLM\footnote{\url{https://github.com/BerriAI/litellm}} as an \emph{LLM proxy} to provided a unified, managed API access to all the individual vLLM instances. Intuitively, LiteLLM behaves as a \emph{reverse proxy}: It exposes an OpenAI-compliant API access point that routes the requests to the corresponding LLM by its name. LiteLLM enables to implement user access control through and usage metering through the use of API Keys. Additionally, we can also use the LLM proxy functionality to seamlessly meter access commercial LLM API providers such as OpenAI and Anthropic, or other research LLM services such as AnvilGPT\footnote{\url{https://www.rcac.purdue.edu/news/6826}}. We issue surrogate API keys to allow us to provide a fine-grained managed access and control budgets. To support Retrieval Augmented generation configurations, we host a Weaviate\footnote{\url{https://github.com/weaviate/weaviate}} vector database environment. Access to Weaviate is controlled too through the use of API keys.
\subsection{Front-end: User facing web interface}
We introduce a web-based software interface designed to complement our model serving and storage features. At this interface, the federated login and authentication is achieved through CILogon\footnote{\url{https://www.cilogon.org/}}. For access control, each individual instructional and research groups typically have a different set of requirements. For example one group might want access to different models while some others might want access only to the vector database or only to the RAG pipeline or various combinations thereof. \verde solves this by customizing a pipeline based on the specific requirements of each group. We implement the abstraction of groups as \emph{courses}. Therefore each course is comprised of a list of students and instructors. All users have access to their individual API key. Instructors also have access to the list of students and to the budget information of the course. 

We also provide a conversational user interface to enable quick access to an LLM, if that is deemed to be the requirement. The LLM can be configured either in \emph{pass through} or \emph{RAG} mode. The former passes the conversation directly to the LLM configured for the course's interface whereas the later executes a RAG workflow using the vector database assigned to the course. The conversational interface manages the conversation's history and persists and retrieves past conversations in a similar fashion to comparable products such as ChatGPT. Further, As an alternative to the conversational user interface, users are able to programmatically access the LLMs using the API keys provided via the web interface. Programmatic access is enabled thanks to the  LiteLLM component and exposes an OpenAI-compliant API interface, which allows the use of popular third-party integration libraries to build AI-powered applications, such as Langchain\footnote{\url{https://www.langchain.com}} and Llamaindex\footnote{\url{https://www.llamaindex.ai}}.

\subsection{Document Intake}
To support RAG workflows, we developed a document intake service that reads various file formats like MSWord, MSPowerPoint slides, PDF files etc., which are then used to generate a corresponding vector database persisted into AI-VERDE's Weaviate service. The document intake service runs independently of the backend and frontend components. Once a vector database is provisioned, it can be configured as part of a course to enable RAG in the conversational UI.

\section{Issues with commercial LLM platforms in an academic setting.}
\label{problems}
 To understand the real-life limitations and challenges of using available LLM-based frameworks (commercial or open source) in academic settings, we first conducted a survey amongst the faculty, staff and students at the University of Arizona (Refer \autoref{apx:survery} for details).  This section outlines and details several of the most prominent concerns identified through this survey, emphasizing the practical and ethical issues that arise when deploying such AI platforms in University campuses. Further, we will discuss how \verde  tackles each of these problems that impede the easy installation and adaptation of LLMs /AI infrastructure on University campuses. \autoref{tab:comparison} provides a comparative visual overview of the features mentioned below, to some of the popular commercial and open source alternates.

\subsection{Intellectual Property, Privacy and Content Ownership}

The primary concern that was raised in the survey, agreed unanimously by all faculty, staff, and students, was the lack of privacy and control over their data. This is because, these platforms often store user data, such as copyrighted course materials or sensitive queries, on corporate servers, where it may be used for downstream model training. This raises issues of data security, intellectual property rights, and confidentiality, particularly in educational settings.

\verde addresses these concerns by ensuring all data, including user queries and uploaded content, is processed entirely on-premises within a secure infrastructure. Unlike commercial platforms, \verde does not store or reuse queries for model training and disables personalization features by default to prioritize privacy. These features are only activated with explicit user consent, maintaining strict control over sensitive academic and personal information.

\subsection{Limited Access to Specialized Knowledge}
General-purpose LLMs platforms, while effective for general information retrieval, often fail to meet domain-specific needs \cite{minaee2024large}. Specialized chatbots for fields like medicine or law are typically limited to answering frequently asked questions and lack the conversational depth required for complex academic queries \cite{yigci2024large}. These limitations reduce accuracy and trust, diminishing their utility in research and education.

As mentioned before, \verde overcomes these challenges by leveraging RAG, an AI methodology that retrieves contextually relevant documents based on user queries and primes the LLM with this information. By indexing specialized knowledge datasets, such as niche research papers or course materials, \verde enables accurate, context-aware responses without storing or training on user-provided data. This ensures data security while empowering researchers and educators to access precise, domain-specific insights.

\begin{table*}[t]
    \centering
    
    \begin{tabular}{lcccc}
    \toprule
    \textbf{\emph{Feature}} & \textbf{\emph{ChatGPT}} & \textbf{\emph{Gemini}} & \textbf{\emph{Anvil}} & \textbf{\emph{\verde}} \\
    \midrule
    Privacy-Preservation & No & No & No & Yes \\
    Built-in guardrails & Yes & Yes & Unknown & Yes \\
    On-Premises Deployment & No& No& No & Yes \\
    Native RAG support & Limited & Limited& No & Yes \\
    Instructional groups management& No& No& No & Yes \\
    Hosted models customization & No& No& No & Yes \\
    \bottomrule
    \end{tabular}
    \caption{Feature comparison between \verde and other LLM comparable platforms. ChatGPT and Gemini are representative of commercial offerings, while Anvil is a representative of an academic offering.}
    \label{tab:comparison}
\end{table*}




\subsection{Authorization and Authentication}
Another major challenge with commercial chatbots in academic settings is their inability to integrate smoothly with existing university platforms, such as authentication and authorization protocols of the respective learning management systems (LMS) \cite{oliveira2016learning}. In the Universities in the United States, for instance, typically proprietary content like grades or course materials in an LMS often requires authentication via dedicated platforms ensuring access is restricted to individuals with university-affiliated email addresses. Furthermore, post-authentication, there are additional authorization requirements, such as ensuring only students registered for a course can access its related chatbot.

Publicly available commercial and open-source platforms typically fail to meet these needs, operating in isolation and lacking flexibility to align with specific university protocols. \verde addresses these concerns by supporting seamless integration with university authentication platforms through CILogon, allowing registered users of University LMS to access services without added complexity. Additionally, \verde simplifies inter-university collaboration by offering a plug-and-play authentication interface, enabling smooth integration and secure cross-institutional collaboration.

\subsection{Equity and Resource Constraints}
Freemium AI platforms often disadvantage financially constrained students and institutions by restricting advanced features to paid subscriptions. Token-based pricing and manual on-boarding processes, even in NSF-funded AnvilGPT\footnote{\url{https://anvilgpt.rcac.purdue.edu/}}, hinder scalability, making it difficult for educators to manage large cohorts. These challenges underscore the need for automated enrollment and management solutions in educational LLM applications while addressing financial compliance concerns in cloud-based resources. Additionally, the increasing reliance on cloud-based resources has introduced significant concerns over budget management and financial compliance for academic departments. 

\verde addresses these challenges through open-source components, RAG technology, and partnerships with cost-effective hardware providers like CyVerse\cite{swetnam2024cyverse} and NSF's Jetstream2 \cite{hancock2021jetstream2}, ensuring minimal operational costs. It also automates budget management by enabling faculty to allocate class-specific funds and distribute API keys efficiently. Moreover, \verde reduces the administrative burden by handling routine queries with chatbots, freeing faculty to focus on research, teaching, and impactful academic work, thus providing a literal egalitarian gateway\footnote{It's egalitarian since any person on a University campus has equal access rights to AI\-VERDE. It is a gateway, since users can access any of their favorite state-of-the-art LLM, picked from the current 
 extremely fast moving LLM production pipeline, while using the same access methodology.}.

\subsection{Anonymity and confidentiality of sensitive data}
Compliance with regulatory standards such as Health Insurance Portability and Accountability Act \cite{act1996health} (HIPAA), Family Educational Rights and Privacy Act \cite{rights2014family} (FERPA), and Data Use Agreements (DUAs) in research and education contexts often prohibits the use of external services for handling sensitive data. In several cases respective institutions have to ensure that external service providers do not inadvertently compromise data privacy or violate agreements of the end-users they cater to. For example, many academic research projects involve human subjects, requiring compliance with strict ethical and legal standards like IRB approval to ensure data privacy and confidentiality. However, the usage of commercial large language model (LLM) platforms raises significant concerns due to ambiguous data handling practices \cite{yao2024survey,wang2023data,jaff2024data}, risking the exposure of sensitive information, including, possibly, Personal Health Information (PHI). Additionally, limiting the tracking of individual user activities is also a priority in certain settings and projects. 

AI-VERDE addresses privacy concerns by operating entirely within a secure, on-premises infrastructure, eliminating risks of external data transfer. It integrates with Soteria\footnote{\url{https://soteria.arizona.edu/}}, a secure data analysis enclave for HIPAA compliance, ensuring secure data processing. Additionally, its gateway model abstracts user interactions, preventing commercial providers from tracking or collecting individual user data, strengthening privacy and supporting ethical data management practices.

\subsection{Access to dedicated hardware}
Effectively harnessing the power of LLMs requires fine-tuning on specialized datasets or integrating them with RAG systems. However, hosting LLMs for a large user base demands dedicated and expensive hardware, with budgets often exceeding \$50,000—a prohibitive expense for most campus researchers (Refer Appendix \ref{cost} for details).

\verde alleviates these challenges by providing a fully equipped hardware infrastructure with pre-loaded LLMs. Further it features elastic hardware allocation tailored to user needs and integrates seamlessly with NSF Jetstream, CyVerse, and cloud services like AWS \footnote{\url{https://aws.amazon.com/}} and Azure \footnote{\url{https://azure.microsoft.com/}}, enabling efficient scaling for diverse research and teaching workloads.

While these are the most prominent problems and their solutions through \verde, there are several more unique contributions of \verde like reduced hallucination, addressing the steep learning curve of AI by providing a dedicated team of consultants and AI educators etc, which are detailed in the Appendix \ref{other_concerns} due to space limitations. 

\section{Pilot Deployment}
Since May 2024, \verde is deployed as part of a pilot project at The University of Arizona. The purpose of this pilot deployment is to stress test the system as well as understand the feasibility of its maintenance while understanding the needs of the academic community, especially in a live implementation context.

\subsection{Qualitative Analysis}

The current users of \verde at the University of Arizona can be divided into three  major user categories.

First, in the education frontier, AI-VERDE serves as a control plane for interacting with both commercial and open-source LLMs. Initially adopted by faculty in AI-related courses like \texttt{INFO 555: Applied Natural Language Processing} and \texttt{BME/SIE 477/577}, it provides features like automated API key generation and budget management, with plans for expansion into non-AI disciplines.

The second major group of AI-VERDE users are researchers who analyze large collections of papers using its LLM-powered capabilities to extract insights and stay updated with fast-paced publications. For example, it was used to index 3,000 papers on antenna methodologies for the Electronics and Communications Engineering Department, enabling efficient retrieval and deeper analysis.

Third category of users were from the support departments of the University. For example, the High-Performance Computing (HPC) department utilizes \verde's RAG-based chatbot to streamline interactions with user documentation, enhancing efficiency and improving the quality of FAQ pages. 

While these are the 3 major categories of clients of \verde, 
there are several others  as shown in Appendix  \ref{supported_fulllist} and \ref{future}.

\subsection{A Quantitative Analysis} During the course of the afore-mentioned pilot deployment of AI-VERDE, 78 different users utilized it for 5 different courses and 10 research projects. The users represent a population of students, instructors, researchers and staff across multiple academic units of the university, \autoref{tab:tokens} shows the token and API consumption statistics of \verde during a period of six months. In brief, 97 thousand API calls representing more than 110 tokens were passed through the variety of models, both self-hosted and relayed to third-party services. While these metrics are relatively low given the potential user base in an institution of this scale, especially since we are still in the pilot mode, the amount of tokens generate represents significant use and its adoption by an increasing user base and hence its huge potential for adoption in academic settings.

\begin{table}[h]
    \centering
    \begin{tabular}{lrrr}
        \toprule
        \multicolumn{4}{c}{\textbf{\textit{Token Consumption}}}\\
        
        & \multicolumn{1}{c}{\emph{Self-hosted}} & \multicolumn{1}{c}{\emph{Proxy}} & \multicolumn{1}{c}{\emph{Total}}\\
        \midrule
         Prompt & 74.96M & .689M & 75.65M \\
         Completion & 34.80M & .230M & 35.03M \\
         Total& 109.76M & .919M & 110.68M \\
         \midrule
         \emph{\textbf{API Calls}} & 96,403 & 1,255 & 97,658\\
         
         \bottomrule
    \end{tabular}
    \caption{\verde usage metrics during the pilot period of 05/30/2024 - 11/26/2024. In the top part of the table, the number of tokens transmitted are shown in millions. Specifically, in rows, prompt tokens represent the input data sent to various models and completion tokens represent the response data generated by the models based on the prompts. In columns, the self-hosted and proxy columns represent tokens fed and generated to models hosted in-premises, and relayed to third-party services such as OpenAI and Anvil, respectively. In the bottom part of the table, API calls, represent the total number of requests made to the models either via the conversational interface and through API access.}
    \label{tab:tokens}
\end{table}
\section{Related Work}
In this section we talk only about prior-art in the field  related specifically to adoption of LLM in University settings ( same for the history of  LLMS can be found in Appendix \ref{related}). Currently after the AI boom,  triggered and accelerated by introduction of ChatGPT \cite{openai2022chatgpt},  several universities, government agencies and private institutions have established collaborations and consortia to reduce the barrier of entry to LLMs for learning groups. Arizona State University (ASU) recently partnered with OpenAI\footnote{\url{https://www.insidehighered.com/news/tech-innovation/artificial-intelligence/2024/05/21/unpacking-asus-openai-partnership-and}} to allow groups to apply for access to ChatGPT for research purposes. Purdue University's Anvil \citep{10.1145/3491418.3530766} provides API access to open-source LLMs through NSF's ACCESS.

Unlike other platforms like Arizona State University’s partnership with OpenAI or Purdue’s Anvil, AI-VERDE offers several additional features, including automated user onboarding and API key management. While on-premises hosting in AI-VERDE mitigates privacy risks, it also provides proxy access to commercial models like OpenAI's GPT. In contrast, Anvil lacks extensive access management, whereas AI-VERDE enhances this by offering tools tailored for instructors to manage user access and allocate budgets to groups and classes.




\section{Conclusion and Future Work}

This work introduces AI-VERDE, a platform providing LLM services tailored for academic environments, prioritizing privacy, accessibility, and adaptability. Developed with open-source components, it offers capabilities similar to commercial LLMs, with features designed for students, faculty, and researchers. By processing all data within institutional infrastructure, AI-VERDE ensures privacy and mitigates external data exposure risks. Its integration with university systems supports research and teaching, fostering an AI-driven ecosystem that promotes innovation, collaboration, and critical thinking in higher education.

In the future, we plan to integrate AI-VERDE into learning management systems like D2L\footnote{\url{https://www.d2l.com}}, enabling course-specific instances where faculty can upload materials and students receive tailored interactions throughout the semester. We also aim to develop AI applications for researchers, such as a user interface for proofreading grant proposals against funding agency requirements and providing feedback to improve submissions.

\section*{Ethics and Limitations}

While \verde is a platform with original source code, it relies exclusively on open source models and software to host and serve LLMs. We stand on the shoulders of the research groups who graciously make their state-of-the-art models available to the general public and of the organizations who release their software freely for non-commercial use. Given that \verde exposes existing LLMs, any inherent biases in those models will be exhibited too by Verde.

One of the principal tenets of our platform is privacy-preservation. While we deliberately don't persist any prompts or responses, we can't extend such guarantee to any group or person who uses \verde as a proxy to manage a budget of an external platform, such as OpenAI. Also while we designed \verde to be a truly egalitarian platform, we do accept that we are still limited by the amount of dedicated hardware available in our deployment, especially from those provided by Cyverse and NSF Jetstream. However, as usage and adoption of \verde increases, possibly outside the University of Arizona itself, we expect applying for new funding to expand our dedicated hardware resources.

\section*{Acknowledgments}

The authors would like to thank Nick Eddy, John Xu, Illyoung Choi, Michele Yung, Mariah Wall, Hagan Franks, Ian Mcewan, Rudy Salcido, Sarah Roberts, Paul Sarando, Maliaca Oxnam, Ajay Perumbeti, Andy Edmonds, Carlos Lizarraga-Celaya, Jeff Gillan, Megh Krishnaswamy, Michele Cosi, Jim Davis, John W, Sean Davey, Tina Johnson, Tina Lee, Tyson Swetnam and Zi K Deng without whose help and support this product would not have happened.

\bibliography{custom}

\begin{thebibliography}{28}
\providecommand{\natexlab}[1]{#1}

\bibitem[{Abdin et~al.(2024)Abdin, Aneja et~al.}]{abdin2024phi}
Marah Abdin, Jyoti Aneja, et~al. 2024.
\newblock Phi-3 technical report: A highly capable language model locally on your phone.
\newblock \emph{arXiv preprint arXiv:2404.14219}.

\bibitem[{Act(1996)}]{act1996health}
Accountability Act. 1996.
\newblock Health insurance portability and accountability act of 1996.
\newblock \emph{Public law}, 104:191.

\bibitem[{Beda et~al.(2014)Beda, Burns, McLuckie et~al.}]{Kubernetes2014}
Joe Beda, Brendan Burns, Craig McLuckie, et~al. 2014.
\newblock \href {https://kubernetes.io/} {Kubernetes: An open-source container orchestration platform}.
\newblock Developed by Google and maintained by the Cloud Native Computing Foundation (CNCF).

\bibitem[{Douze et~al.(2024)Douze, Guzhva, Deng, Johnson, Szilvasy, Mazaré, Lomeli, Hosseini, and Jégou}]{douze2024faiss}
Matthijs Douze, Alexandr Guzhva, Chengqi Deng, Jeff Johnson, Gergely Szilvasy, Pierre-Emmanuel Mazaré, Maria Lomeli, Lucas Hosseini, and Hervé Jégou. 2024.
\newblock \href {https://arxiv.org/abs/2401.08281} {The faiss library}.

\bibitem[{Dubey et~al.(2024)Dubey, Jauhri, Pandey, Kadian, Al-Dahle, Letman, Mathur, Schelten, Yang, Fan et~al.}]{dubey2024llama}
Abhimanyu Dubey, Abhinav Jauhri, Abhinav Pandey, Abhishek Kadian, Ahmad Al-Dahle, Aiesha Letman, Akhil Mathur, Alan Schelten, Amy Yang, Angela Fan, et~al. 2024.
\newblock The llama 3 herd of models.
\newblock \emph{arXiv preprint arXiv:2407.21783}.

\bibitem[{Hancock et~al.(2021)Hancock, Fischer, Lowe, Snapp-Childs, Pierce, Marru, Coulter, Vaughn, Beck, Merchant et~al.}]{hancock2021jetstream2}
David~Y Hancock, Jeremy Fischer, John~Michael Lowe, Winona Snapp-Childs, Marlon Pierce, Suresh Marru, J~Eric Coulter, Matthew Vaughn, Brian Beck, Nirav Merchant, et~al. 2021.
\newblock Jetstream2: Accelerating cloud computing via jetstream.
\newblock In \emph{Practice and Experience in Advanced Research Computing}, pages 1--8.

\bibitem[{Hu et~al.(2022)Hu, Shen, Wallis, Allen-Zhu, Li, Wang, Wang, and Chen}]{hu2022lora}
Edward~J Hu, Yelong Shen, Phillip Wallis, Zeyuan Allen-Zhu, Yuanzhi Li, Shean Wang, Lu~Wang, and Weizhu Chen. 2022.
\newblock \href {https://openreview.net/forum?id=nZeVKeeFYf9} {Lo{RA}: Low-rank adaptation of large language models}.
\newblock In \emph{International Conference on Learning Representations}.

\bibitem[{Jaff et~al.(2024)Jaff, Wu, Zhang, and Iqbal}]{jaff2024data}
Evin Jaff, Yuhao Wu, Ning Zhang, and Umar Iqbal. 2024.
\newblock Data exposure from llm apps: An in-depth investigation of openai's gpts.
\newblock \emph{arXiv preprint arXiv:2408.13247}.

\bibitem[{Jiang et~al.(2023{\natexlab{a}})Jiang, Sablayrolles, Mensch, Bamford, Chaplot, Casas, Bressand, Lengyel, Lample, Saulnier et~al.}]{jiang2023mistral}
Albert~Q Jiang, Alexandre Sablayrolles, Arthur Mensch, Chris Bamford, Devendra~Singh Chaplot, Diego de~las Casas, Florian Bressand, Gianna Lengyel, Guillaume Lample, Lucile Saulnier, et~al. 2023{\natexlab{a}}.
\newblock Mistral 7b.
\newblock \emph{arXiv preprint arXiv:2310.06825}.

\bibitem[{Jiang et~al.(2023{\natexlab{b}})Jiang, Sablayrolles, Mensch, Bamford, Chaplot, de~las Casas, Bressand, Lengyel, Lample, Saulnier, Lavaud, Lachaux, Stock, Scao, Lavril, Wang, Lacroix, and Sayed}]{jiang2023mistral7b}
Albert~Q. Jiang, Alexandre Sablayrolles, Arthur Mensch, Chris Bamford, Devendra~Singh Chaplot, Diego de~las Casas, Florian Bressand, Gianna Lengyel, Guillaume Lample, Lucile Saulnier, Lélio~Renard Lavaud, Marie-Anne Lachaux, Pierre Stock, Teven~Le Scao, Thibaut Lavril, Thomas Wang, Timothée Lacroix, and William~El Sayed. 2023{\natexlab{b}}.
\newblock \href {https://arxiv.org/abs/2310.06825} {Mistral 7b}.
\newblock \emph{Preprint}, arXiv:2310.06825.

\bibitem[{Kwon et~al.(2023)Kwon, Li, Zhuang, Sheng, Zheng, Yu, Gonzalez, Zhang, and Stoica}]{kwon2023efficient}
Woosuk Kwon, Zhuohan Li, Siyuan Zhuang, Ying Sheng, Lianmin Zheng, Cody~Hao Yu, Joseph~E. Gonzalez, Hao Zhang, and Ion Stoica. 2023.
\newblock Efficient memory management for large language model serving with pagedattention.
\newblock In \emph{Proceedings of the ACM SIGOPS 29th Symposium on Operating Systems Principles}.

\bibitem[{Lewis et~al.(2020)Lewis, Perez, Piktus, Petroni, Karpukhin, Goyal, K\"{u}ttler, Lewis, Yih, Rockt\"{a}schel, Riedel, and Kiela}]{NEURIPS2020_6b493230}
Patrick Lewis, Ethan Perez, Aleksandra Piktus, Fabio Petroni, Vladimir Karpukhin, Naman Goyal, Heinrich K\"{u}ttler, Mike Lewis, Wen-tau Yih, Tim Rockt\"{a}schel, Sebastian Riedel, and Douwe Kiela. 2020.
\newblock \href {https://proceedings.neurips.cc/paper_files/paper/2020/file/6b493230205f780e1bc26945df7481e5-Paper.pdf} {Retrieval-augmented generation for knowledge-intensive nlp tasks}.
\newblock In \emph{Advances in Neural Information Processing Systems}, volume~33, pages 9459--9474. Curran Associates, Inc.

\bibitem[{Minaee et~al.(2024)Minaee, Mikolov, Nikzad, Chenaghlu, Socher, Amatriain, and Gao}]{minaee2024large}
Shervin Minaee, Tomas Mikolov, Narjes Nikzad, Meysam Chenaghlu, Richard Socher, Xavier Amatriain, and Jianfeng Gao. 2024.
\newblock Large language models: A survey.
\newblock \emph{arXiv preprint arXiv:2402.06196}.

\bibitem[{Oliveira et~al.(2016)Oliveira, Cunha, and Nakayama}]{oliveira2016learning}
Paulo Cristiano~de Oliveira, Cristiano Jose Castro de~Almeida Cunha, and Marina~Keiko Nakayama. 2016.
\newblock Learning management systems (lms) and e-learning management: an integrative review and research agenda.
\newblock \emph{JISTEM-Journal of Information Systems and Technology Management}, 13(2):157--180.

\bibitem[{OpenAI(2023)}]{openai2023gpt4}
OpenAI. 2023.
\newblock \href {https://arxiv.org/abs/2303.08774} {Gpt-4 technical report}.
\newblock \emph{arXiv preprint arXiv:2303.08774}.

\bibitem[{OpenAI(2022)}]{openai2022chatgpt}
TB~OpenAI. 2022.
\newblock Chatgpt: Optimizing language models for dialogue. openai.

\bibitem[{Reimers and Gurevych(2019)}]{reimers-2019-sentence-bert}
Nils Reimers and Iryna Gurevych. 2019.
\newblock \href {https://arxiv.org/abs/1908.10084} {Sentence-bert: Sentence embeddings using siamese bert-networks}.
\newblock In \emph{Proceedings of the 2019 Conference on Empirical Methods in Natural Language Processing}. Association for Computational Linguistics.

\bibitem[{Rights and Act(2014)}]{rights2014family}
Family~Educational Rights and Privacy Act. 2014.
\newblock Family educational rights and privacy act (ferpa).

\bibitem[{Simon(2021)}]{Simon_2021}
Julien Simon. 2021.
\newblock \href {https://huggingface.co/blog/large-language-models} {Large language models: A new moore’s law?}

\bibitem[{Song et~al.(2022)Song, Smith, Kalyanam, Zhu, Adams, Colby, Finnegan, Gough, Hillery, Irvine, Maji, and St.~John}]{10.1145/3491418.3530766}
X.~Carol Song, Preston Smith, Rajesh Kalyanam, Xiao Zhu, Eric Adams, Kevin Colby, Patrick Finnegan, Erik Gough, Elizabett Hillery, Rick Irvine, Amiya Maji, and Jason St.~John. 2022.
\newblock \href {https://doi.org/10.1145/3491418.3530766} {Anvil - system architecture and experiences from deployment and early user operations}.
\newblock In \emph{Practice and Experience in Advanced Research Computing 2022: Revolutionary: Computing, Connections, You}, PEARC '22, New York, NY, USA. Association for Computing Machinery.

\bibitem[{Swetnam et~al.(2024)Swetnam, Antin, Bartelme, Bucksch, Camhy, Chism, Choi, Cooksey, Cosi, Cowen et~al.}]{swetnam2024cyverse}
Tyson~L Swetnam, Parker~B Antin, Ryan Bartelme, Alexander Bucksch, David Camhy, Greg Chism, Illyoung Choi, Amanda~M Cooksey, Michele Cosi, Cindy Cowen, et~al. 2024.
\newblock Cyverse: Cyberinfrastructure for open science.
\newblock \emph{PLoS Computational Biology}, 20(2):e1011270.

\bibitem[{Touvron et~al.(2023)Touvron, Martin, Stone, Albert, Almahairi, Babaei, Bashlykov, Batra, Bhargava, Bhosale, Bikel, Blecher, Ferrer, Chen, Cucurull, Esiobu, Fernandes, Fu, Fu, Fuller, Gao, Goswami, Goyal, Hartshorn, Hosseini, Hou, Inan, Kardas, Kerkez, Khabsa, Kloumann, Korenev, Koura, Lachaux, Lavril, Lee, Liskovich, Lu, Mao, Martinet, Mihaylov, Mishra, Molybog, Nie, Poulton, Reizenstein, Rungta, Saladi, Schelten, Silva, Smith, Subramanian, Tan, Tang, Taylor, Williams, Kuan, Xu, Yan, Zarov, Zhang, Fan, Kambadur, Narang, Rodriguez, Stojnic, Edunov, and Scialom}]{touvron2023llama2openfoundation}
Hugo Touvron, Louis Martin, Kevin Stone, Peter Albert, Amjad Almahairi, Yasmine Babaei, Nikolay Bashlykov, Soumya Batra, Prajjwal Bhargava, Shruti Bhosale, Dan Bikel, Lukas Blecher, Cristian~Canton Ferrer, Moya Chen, Guillem Cucurull, David Esiobu, Jude Fernandes, Jeremy Fu, Wenyin Fu, Brian Fuller, Cynthia Gao, Vedanuj Goswami, Naman Goyal, Anthony Hartshorn, Saghar Hosseini, Rui Hou, Hakan Inan, Marcin Kardas, Viktor Kerkez, Madian Khabsa, Isabel Kloumann, Artem Korenev, Punit~Singh Koura, Marie-Anne Lachaux, Thibaut Lavril, Jenya Lee, Diana Liskovich, Yinghai Lu, Yuning Mao, Xavier Martinet, Todor Mihaylov, Pushkar Mishra, Igor Molybog, Yixin Nie, Andrew Poulton, Jeremy Reizenstein, Rashi Rungta, Kalyan Saladi, Alan Schelten, Ruan Silva, Eric~Michael Smith, Ranjan Subramanian, Xiaoqing~Ellen Tan, Binh Tang, Ross Taylor, Adina Williams, Jian~Xiang Kuan, Puxin Xu, Zheng Yan, Iliyan Zarov, Yuchen Zhang, Angela Fan, Melanie Kambadur, Sharan Narang, Aurelien Rodriguez, Robert Stojnic, Sergey Edunov, and Thomas Scialom. 2023.
\newblock \href {https://arxiv.org/abs/2307.09288} {Llama 2: Open foundation and fine-tuned chat models}.
\newblock \emph{Preprint}, arXiv:2307.09288.

\bibitem[{Vaswani et~al.(2017)Vaswani, Shazeer, Parmar, Uszkoreit, Jones, Gomez, Kaiser, and Polosukhin}]{NIPS2017_3f5ee243}
Ashish Vaswani, Noam Shazeer, Niki Parmar, Jakob Uszkoreit, Llion Jones, Aidan~N Gomez, \L~ukasz Kaiser, and Illia Polosukhin. 2017.
\newblock \href {https://proceedings.neurips.cc/paper_files/paper/2017/file/3f5ee243547dee91fbd053c1c4a845aa-Paper.pdf} {Attention is all you need}.
\newblock In \emph{Advances in Neural Information Processing Systems}, volume~30. Curran Associates, Inc.

\bibitem[{Wang et~al.(2023{\natexlab{a}})Wang, Mao, Wu, Ge, Wei, and Ji}]{wang2023unleashing}
Zhenhailong Wang, Shaoguang Mao, Wenshan Wu, Tao Ge, Furu Wei, and Heng Ji. 2023{\natexlab{a}}.
\newblock Unleashing the emergent cognitive synergy in large language models: A task-solving agent through multi-persona self-collaboration.
\newblock \emph{arXiv preprint arXiv:2307.05300}.

\bibitem[{Wang et~al.(2023{\natexlab{b}})Wang, Zhong, Wang, Zhu, Mi, Wang, Shang, Jiang, and Liu}]{wang2023data}
Zige Wang, Wanjun Zhong, Yufei Wang, Qi~Zhu, Fei Mi, Baojun Wang, Lifeng Shang, Xin Jiang, and Qun Liu. 2023{\natexlab{b}}.
\newblock Data management for large language models: A survey.
\newblock \emph{CoRR}.

\bibitem[{Yao et~al.(2024)Yao, Duan, Xu, Cai, Sun, and Zhang}]{yao2024survey}
Yifan Yao, Jinhao Duan, Kaidi Xu, Yuanfang Cai, Zhibo Sun, and Yue Zhang. 2024.
\newblock A survey on large language model (llm) security and privacy: The good, the bad, and the ugly.
\newblock \emph{High-Confidence Computing}, page 100211.

\bibitem[{Yigci et~al.(2024)Yigci, Eryilmaz, Yetisen, Tasoglu, and Ozcan}]{yigci2024large}
Defne Yigci, Merve Eryilmaz, Ail~K Yetisen, Savas Tasoglu, and Aydogan Ozcan. 2024.
\newblock Large language model-based chatbots in higher education.
\newblock \emph{Advanced Intelligent Systems}, page 2400429.

\bibitem[{Zhou et~al.(2022)Zhou, Muresanu, Han, Paster, Pitis, Chan, and Ba}]{zhou2022large}
Yongchao Zhou, Andrei~Ioan Muresanu, Ziwen Han, Keiran Paster, Silviu Pitis, Harris Chan, and Jimmy Ba. 2022.
\newblock Large language models are human-level prompt engineers.
\newblock \emph{arXiv preprint arXiv:2211.01910}.

\end{thebibliography}
\bibliographystyle{acl_natbib}
\newpage

\appendix

\textbf{Appendix}

\section{Survey details}
\label{apx:survery}

 To understand the pulse of faculty, student and staff in a university campus towards adoption of AI for education and research purposes, a  survey was conducted in Spring 2024, at he University of Arizona. It must be noted that this survey deviated from the standard modality of a single survey comprising various questions, that would be filled by all users. Instead it was done in several different modalities based on various factors including count and convenience of the intended group. For example with Faculty, it was easier to do comprehensive hour long one on one interviews. While with undergrad students, since the real estate to cover was larger, focus groups were conducted, where the students were asked to debate about this topic.

 In all the 3 groups, first a separation between participants who are for and against AI was made. Then from within the latter sub sample, various modalities of information extraction was done, as detailed below.

In case of  undergraduate students a flyer was first sent out asking for students who are interested in AI adoption on campus, to sign up. From all the applicants who signed up, a sub sample of 53 students were picked up. This sampling ensured that it included a fair representation of students from different levels and modes of undergraduate journey, including, freshmen through seniors,  students with multiple majors, spread across several different colleges etc. This ensured that the final sample of students picked represented a fair representation of every undergraduate persona groups distributed across campus. The final 53 students were further divided into five groups. Each of these groups were assembled in person, and a group discussion was conducted, each an hour long. At the end of the hour they were asked to come up with their top 10 concerns regarding adoption of AI on college campus.  .

Meanwhile with the graduate students, since the sample size was lesser, eight in depth one- on-one interviews were conducted. Further, 27 others responded to email survey question on AI needs and interests. Like the undergraduate sample, it was ensured that this represented graduate students across various colleges and disciplines, various levels of their journey, including masters and Phd students.

Amongst the faculty and staff of University of Arizona,
41 one-on-one interviews were conducted to understand in-depth their needs and wish list with AI tools, in supporting their research and instructional needs. Do note that starting sample was less than 5, but these snowballed into referrals to other faculty who they thought would be definitely interested in responding to these. These one-on-one interviews included 12 research faculty and 2 UA library staff who were already using AI tools.

Apart from all these efforts,  an email survey with open-ended questions was sent out to 112 faculty, staff, and graduate students who were part of the  Faculty Learning Community (FLC)\footnote{FLC is a peer-led group of faculty members (typically 6 to 12) who engage in an active, collaborative, year–long program, structured to provide encouragement, support, and reflection in teaching and learning.} at University of Arizona.  From this group we received 57 responses.

Further there was an open call for AI needs, ideas and scenarios that went s at University of Arizona like the AI\^2 Task Force \footnote{\url{https://artificialintelligence.arizona.edu/about-us}}, University Center for Assessment, Teaching and Technology
\footnote{\url{https://ucatt.arizona.edu/}}, and to the staff of University of Arizona Libraries \footnote{\url{https://library.arizona.edu/}}, and some to non-AI focused FLCs like Agriculture, Life and Veterinary Sciences, and Cooperative Extension\footnote{\url{https://alvsce.arizona.edu/}} and CALES \footnote{\url{https://research.cales.arizona.edu/}}. This yield 194 responses. So in total, 372 participants from various walks of life in our University campus were interviewed for this survey. In addition, document analysis  and prior art search was done to identify potential uses and the current state of private or public LLMs in Higher Education.

\section{Other Concerns and solutions offered thereof by \verde}\label{other_concerns}

In this section we detail some more of the concerns raised in the survey, and how \verde solves these issues. 

\subsection{Steep Learning Curve}
The primary uses of LLMs on university campuses are supporting teaching and enhancing research. While faculty and researchers are the main users, they often face challenges due to a lack of programming or AI expertise, particularly those from non-STEM disciplines. Current LLM solutions require significant technical skills for fine-tuning on specialized data or implementing RAG, making them inaccessible to many.

\verde addresses these barriers by offering tailored support, including workshops for users at all skill levels and personalized planning sessions with AI specialists. This helps non-technical users adopt AI while enabling advanced users to leverage APIs and microservices for tasks like fine-tuning and RAG. Through these initiatives, \verde democratizes AI use, reducing the steep learning curve associated with LLM adoption.

Further, recognizing that AI adoption varies across user needs, \verde provides personalized planning sessions with AI specialist consultants. These consultants assess individual requirements—whether for researchers using personal GPUs or educators integrating AI into teaching—and design tailored solutions to optimize resource use. They also guide users through integration and provide hands-on training, significantly lowering barriers to AI adoption.

Also for experienced users with programming knowledge, \verde offers direct API access and microservices, enabling advanced tasks such as fine-tuning and RAG. Additionally, workshops and training sessions empower technical users to maximize the platform's capabilities. For example some users (e.g researchers on campus) would like to use their Laptop or their own GPU and not use a high performance computing or Cyverse. So to support it, AI Verde meets the person where they are. Which is why the very first step in AI Verde will be a conversation with our AI Specialist consultants who will give you a plan and path on how best you can use the available resources.

To summarize from the experience of building \verde we learned that AI/LLM offering, is not a one-size-fits-all-solution and hence we support such varying levels of user skills.

\subsection{Hallucination and Misinformation}

Another key concern highlighted in the survey is the issue of hallucination, where general-purpose chatbots produce responses that are plausible but factually incorrect. Research has shown that this happens when chatbots struggle to find answers to questions. This issue is particularly problematic in academic settings, as students may unknowingly rely on and learn from incorrect information, leading to negative learning outcomes. The issue is further complicated by broader ethical concerns such as enabling cheating, plagiarism, and breaches of academic integrity standards.

To reduce hallucination,  \verde utilizes advanced prompt engineering techniques. Through extensive trials, a prompt was developed to ensure the platform responds with a clear negative acknowledgment (e.g.,"That question is beyond my purview of current knowledge") when it cannot provide a correct answer, as opposed to hallucination. While we don't claim that the problem of hallucination has been completely solved, but when combined with appropriate temperature settings and RAG, \verde grounds its responses in verified, retrieved documents. This approach ensures that the generated outputs are accurate, reliable, and aligned with the original source material with minimal hallucination .

\subsection{Guard rails}
Another major concern raised in the survey was the possibilty of inappropriate language used, either by the user or by the LLM in its response. To address this \verde incorporates robust solutions  like Llama guardrails \footnote{\url{https://www.llama.com/docs/model-cards-and-prompt-formats/llama-guard-3/}} to prevent biased or inappropriate outputs. 
\subsection{Prompt Engineering} \label{prompt_eng}

Effective utilization of LLMs require careful prompt engineering \cite{zhou2022large,wang2023unleashing}. Note that prompt engineering doesn't need programming or AI knowledge, so can be done by a Professor or Researcher with minimal training. However, prompt engineering gets trickier within a RAG system. That is because  the prompt supplied by the user on a typical chatbot based interfaced, is used to query relevant documents from the vector database. However, the actual prompt presented to the LLM is different and the end-user doesn't have access to it. This lack of access to the final prompt restricts the end user. 

To overcome this \verde provides API and Chatbot based access to most of the state-of-the-art LLMs (both the paid commercial ones and the free open source ones), builds a RAG pipeline over a specialized vector database created from specialized knowledge provided by an end user. In fact chatbot is only one of the services provided by the platform. Every module of the platform is exposed as a micro-service for the end-user to experiment with. Thus for a researcher who wants to explore prompt engineering with RAG, we give them ability to create and access the vector database, and a subsequent access to modify system prompt using programmatic access. For example here is a prompt created by the researcher of Antennas who wanted deeper inferences, and was ready to pass custom documents that were not in the vector database. ``You are a teaching assistant. 
You are having a conversation with a student and the student will ask you a question. 
To answer the student's question use information only from the reference text that 
is between <Reference></Reference> and from the history of the conversation. When you answer the question, quote the text that you used to base your answer off. If you can't answer it, then say "I can't answer this question". You will add the URL for the source if it is available.
You always answer the question with markdown formatting. 
You will be penalized if you do not answer with markdown when it would be possible. The markdown formatting you support: headings, bold, italic, links, tables, lists, code blocks, and blockquotes. You do not support images and never include images. You will be penalized if you render images. You will not wrap the output with triple backticks.
Reference text:<Reference><Reference>""

\subsection{Reducing the Entry barrier to AI}

While AI and LLM might not be that intimidating to the STEM majors, for a student or faculty coming from disciplines like humanities and social sciences even python programming will be intimidating, let alone details of AI. Note that not everything in AI Verde needs programming knowledge- everything is based on the need of the end user. AI Verde understands this, and like we mentioned above , would like to meet the person where they are. Hence we provide training workshops starting from programming workshops all the way through advanced AI. Like we mentioned it is not just a software providing platform, but human consultations are the very first step we offer, understanding where exactly the user stands, in a compassionate and no judgment manner. Having said that, our goal is not to give fish to everyone who wants fish, but instead teach them how to fish. Vision and mission of AI Verde (and in turn the Data Science Institute at University of Arizona, which created \verde) is to enable the initial hand holding and pointers required towards making you successful in using the state of the art AI for your researching and teaching resources.

\subsection{More Unique Contributions of \verde} \label{unique_contribs}
As mentioned earlier, while there are the 3 major categories of clients of\verde, 
there are several other use cases, like enabling access to OpenAI-compatible tools (e.g., ChatboxAI\footnote{\url{https://chatboxai.app/en}}) , that highlight \verde's versatile design, and ecosystem compatibility. Further, to address gaps in institutional policies, \verde collaborated with the School of Information to develop a policy framework  for API access, offering a scalable model for other institutions \footnote{This document is provided as supplementary material, so that it can act as a model for other institutions contemplating on similar ventures through \verde or otherwise.}. Also its integration with CILogon ensures compliance by restricting access to authorized users while removing logistical burdens, allowing educators and researchers to focus on academic content without managing LLM provisioning. This vision supports equitable and efficient LLM access for the entire university community. 

The key challenge in provisioning LLM based access for students, courses, workshops etc. is onboarding or enrolling participants and often deprovisioning them when event concludes. This all is done manually and one by one. There are no ready made solutions that handle these. There are no known solutions. Even for NSF funded AnvilGPT each user has to create individual account and the onboarding will take at best 24-48 hours and for educators ensuring a class of 50 students will do this is a huge burden and hurdle even if the resource is free/no cost
We have to comply by HIPAA, FERPA and Data User Agreements (DUA) that often forbid use of external services. We also want to limit how our users usage is tracked so the gateway model provides that abstraction and commercial provides cannot track individuals
Maintaining budgets per course/team and per person is important for departments to ensure financial compliance i.e cloud bills are concern
Academic computing departments at universities are not equipped to handle these new requirements as yet, we contacted universities that have licensed OpenAI and these management functions as absent .
There is a large ecosystem of tools like desktop apps and code assistants that users like to use, VERDE API is OpenAI compatible and things work out of the box. Further until \verde there were no institutional use policies for API access. We developed one had have provided herewith. As institutions we have to support the amazing ecosystem of commercial and on premise capabilities to stay in compliance. Our vision is to give the equivalent of digital birth right for every U of A person to have LLM access to learn, experiment and build. When they need higher capacity/capability as part of their course or research project they will get that through the course or lab. PI's can bring their entitlements from other sources e.g. NSF ACCESS, NAIR and manage the access and budgets. With direct institutional log in integration via CILogon we can ensure that resources access is limited to the individual and meets institutional compliance e.g. cannot access VERDE resources once they do not have institutional affiliation
By removing the provisioning burden and not choosing favorite LLM providers educators and researchers can focus on their material and not have to be burdened with managing LLM access

Overall, \verde strives to provide integrate the underpinning technologies required to develop AI applications using open source software and models, and provide a seamless platform that caters to educators and researchers within an academic environment such that they can focus on developing instructional experiences and research projects without the burden of provisioning, deployment and management.

\subsection{Some other concerns}

Apart from these major concerns some minor concerns raised included that chatbots currently lack the advanced  reasoning ability to evaluate subjective assignments or provide meaningful feedback tied to learning outcomes. Further, it was mentioned that the chatbots are minimizing personal connection i.e., he automated interactions from chat  bots undermine the forming of supportive instructor-student relationships and nuanced communications.  Poor adaptability was another reason. For example, rigid chatbot capabilities cannot readily adapt coaching, guidance,  and support to individual student needs and challenges. Any misunderstandings or errors from the chatbot  on student inquiries or input data undermine its credibility as a knowledge source.

Note that in its present form \verde does not address these challenges but is definitely part of the planned future work.

\section{Full client list}
\label{supported_fulllist}
These are the remaining list of courses and labs currently being supported by \verde at the University of Arizona. Specifically \verde has chat interfaces trained to reply questions related to : 

\begin{enumerate}
    \item Course content for RNR355: Introduction to Wildland Fire.
    \item Research publications from the University of Arizona's Cooperative Extension \footnote{\url{https://extension.arizona.edu/}}.
    \item Content from website and documentation for CyVerse: A computational framework designed to handle large datasets and complex analyses.
    \item  Content from website and documentation for Tech Launch Arizona \footnote{\url{https://techlaunch.arizona.edu/}}: Facilitating the commercialization of University of Arizona inventions.
    
    \item  Content from website and documentation MKDocs \footnote{\url{https://www.mkdocs.org/}}: A static site generator for documentation projects.

    \item  Content from website and documentation MKDocs \footnote{\url{https://www.chishiki-ai.org/}}: an AI-powered Civil Engineering Community.

    \item  Content and  publications from Harwood Lab" \footnote{\url{https://comm.arizona.edu/person/jake-harwood}}

    \item  Content and  publications from Harwood Lab" \footnote{\url{https://comm.arizona.edu/person/jake-harwood}}

    \item  Content and  publications from Harwood Lab" \footnote{\url{https://comm.arizona.edu/person/joseph-bonito}}

    \item  Content and  publications from Eller Partnership Office: \footnote{\url{https://eller.arizona.edu/engage/partnerships-office}}

 \item  Content and  publications related to Antenna research for Hao Xin lab: \footnote{\url{https://ece.engineering.arizona.edu/faculty-staff/faculty/hao-xin}}

    \item  Provides access to AnvilGPT models \footnote{\url{https://anvilgpt.rcac.purdue.edu/}}

\item  Course content for INFO 523 2024 Fall : Data Mining and Discovery.

\end{enumerate}

\section{Some more potential and future use cases} \label{future}

In this section we detail some more future work, which are either planned or on which already development is being done as a feature addition to \verde. 
\subsection{Visibility into Course Effectiveness} By serving as a chatbot assisting students with course material, \verde reduces the instructor's workload, particularly on platforms like Piazza. Faculty can gain valuable insights into the types of questions students are asking, helping assess the class's overall understanding of the material.
\subsection{Instructor Feedback on Teaching Methods} \verde provides indirect feedback on teaching methodologies by tracking student interactions. This data helps faculty evaluate whether their teaching approach is achieving the desired learning outcomes.

\subsection{Teaching Assistant}
In academia, the complexity of coursework increases significantly as students progress through their studies. \verde serves as a helpful assistant, providing students with accurate, domain-specific answers drawn from course materials. Unlike general-purpose chatbots, \verde is designed to provide tailored responses based on specific course content. In its initial release, \verde is trained to answer questions from five diverse datasets, supporting students in a wide range of academic disciplines.

\subsection{Academic and Research Computing Infrastructure Providers}

\subsection{Information Technology Support}

\verde can also enhance the operations of information technology based support departments. For example, the University's Information Technology Services can use \verde to build a classification and redirection system for support tickets. \verde can filter support requests, automatically categorizing them or suggesting solutions based on existing FAQs. This reduces the workload of support staff by automating the triage process, ensuring that only tickets requiring human attention are escalated.

\section {Discussion}
 Currently, it has become inevitable that students are using (and will continue to use) LLM-based systems and chatbots  to assist them in learning and even abetting them in doing their homework assignments. Hence, one of  the goals of designing AI Verde was to eventually become an  alternative to the hallucinating commercial or publicly available options. For example, instead of letting students rely on and learn wrong facts from a constantly hallucinating  universal chatbot like ChatGPT, AI Verde can help them learn from our own custom-developed, well sand-   boxed, private chatbot, specifically trained on only the respective course materials. 
 With AI Verde, we hope to open another front of innovation in pedagogy. We hope faculty in Universities   will consider this as an opportunity to adapt to the ‘novus mundus’ of AI assisted learning. For example,  with AI Verde in place, the instructor, when giving out homework can now suggest: here are the answers   that AI Verde will tell you for this question, but your goal is to beat AI Verde or do better than AI Verde.  
 In summary, what we are trying to achieve through AI Verde and his future progeny, is to build the Ship of Theseus for AI technologies focusing on its use for our community. To quote what an AI researcher  recently said, “AI is not a done deal. We are building the road as we walk it, and we can collectively decide  what direction we want to go in, together.” We think those are really wise words, and we hope that we can  build an AI that really is good for humans, and not necessarily for machines themselves.

\section{Cost details} 
\label{cost}
 Hosting LLMs for a large user base demands dedicated and expensive hardware, with budgets often exceeding \$50,000—a prohibitive expense for most campus researchers.
\section{Related Work}\label{related}

LLMs have proliferated quickly since the inception of the transformers architecture \citep{NIPS2017_3f5ee243}. As the size of these models scaled very quickly \citep{Simon_2021}, the increased computing requirements have driven users towards deploying modern LLMs in client-server environments. As a result, multiple open source software projects designed to efficiently serve models have emerged\footnote{\url{https://ollama.com}}\footnote{\url{https://huggingface.co/docs/text-generation-inference/en/index}}\footnote{\url{https://github.com/vllm-project/vllm}}. These software systems model their application programming interfaces (API) after OpenAI's proprietary API, which allows increases the interoperability and modularity in software development. LLM gateway software such as LiteLLM leverages this interoperability to expose a unified API proxy service that manages access and usage to multiple models. At the heart of \verde, we integrate these technologies to take away the burden of orchestrating and managing the provisioning of open source LLMs.

LLMs are very often used in tandem with external information repositories and databases to build \emph{retrieval augmented generation} (RAG) applications \citep{NEURIPS2020_6b493230}. Often, the information is stored as documents, which are encoded using transformer models designed to compute pair-wise semantic similarity \citep{reimers-2019-sentence-bert}. For this particular use case, multiple vector database management solutions exist that leverage optimized algorithms to compute exact or approximate nearest-neighbors search \citep{douze2024faiss}. \verde integrates as a central element of its platform to support the development of RAG applications in the academic setting.

\end{document}